\documentclass[pdflatex, sn-mathphys-num]{sn-jnl}% Math and Physical Sciences Numbered Reference Style
\usepackage{graphicx}%
\usepackage{multirow}%
\usepackage{amsmath,amssymb,amsfonts}%
\usepackage{amsthm}%
\usepackage{mathrsfs}%
\usepackage[title]{appendix}%
\usepackage{xcolor}%
\usepackage{textcomp}%
\usepackage{manyfoot}%
\usepackage{booktabs}%
\usepackage{algorithm}%
\usepackage{algorithmicx}%
\usepackage{algpseudocode}%
\usepackage{listings}%
\usepackage{macros}

\theoremstyle{thmstyleone}%
\theoremstyle{thmstyletwo}%

\theoremstyle{thmstylethree}%

\begin{document}

\title[]{\ours: Vision-Based Pose Optimization with Fine-Tuned Foundation Models for Accurate Bronchoscopy Navigation}
%BronchOpt: Representation-Guided Pose Optimization for Accurate 2D–3D Registration in Bronchoscopy Navigation


\author*[1]{\fnm{Hongchao} \sur{Shu}}\email{hshu4@jhu.edu}
\author[1]{\fnm{Roger D.} \sur{Soberanis-Mukul}}
\author[1]{\fnm{Jiru} \sur{Xu}}
\author[1]{\fnm{Hao} \sur{Ding}}
\author[2]{\fnm{Morgan} \sur{Ringel}}
\author[2]{\fnm{Mali} \sur{Shen}}
\author[2]{\fnm{Saif Iftekar} \sur{Sayed}}
\author[2]{\fnm{Hedyeh} \sur{Rafii-Tari}}
\author*[1]{\fnm{Mathias} \sur{Unberath}}\email{unberath@jhu.edu}
\affil*[1]{\orgname{Johns Hopkins Univeristy}, \orgaddress{\city{Baltimore}, \state{Maryland}, \postcode{21218}, \country{USA}}}

\affil[2]{\orgdiv{Johnson \& Johnson MedTech}, \orgaddress{\city{Santa Clara}, \state{California}, \postcode{95054},  \country{USA}}}


\abstract{\textbf{Purpose:} Accurate intra-operative localization of the endoscope tip relative to the anatomy remains a major challenge in bronchoscopy due to respiratory motion, anatomical variability, and CT-to-body divergence, which cause deformation and misalignment between intra-operative views and pre-operative CT. Existing vision-based methods often struggle generalize across domains and patients, limiting robustness and leaving residual alignment errors. This work aims to establish a generalizable foundation for bronchoscopy navigation through a robust vision-based framework and a new synthetic benchmark dataset that enables standardized evaluation and reproducible development.

\textbf{Methods:} We propose a vision-based pose optimization framework for frame-wise 2D–3D registration between intra-operative endoscopic views and pre-operative CT anatomy. A fine-tuned modality- and domain-invariant encoder enables direct similarity measurements between real endoscopic RGB images and CT-rendered depth maps, while differentiable rendering refines camera poses through depth consistency. To enhance reproducibility, we introduce the first public synthetic benchmark dataset for bronchoscopy navigation to address the lack of publicly available paired CT–endoscopy data.

\textbf{Results:} Trained solely on synthetic data distinct from the benchmark, our model attains an average translational error of $2.65$ mm and a rotational error of $0.19$ rad, demonstrating high localization accuracy and stability.
Qualitative results on real patient data further confirm strong cross-domain generalization, achieving consistent frame-wise 2D–3D alignment without domain-specific adaptation.

\textbf{Conclusion:} The proposed framework achieves robust, domain-invariant bronchoscopy localization through iterative vision-based optimization, offering a scalable solution toward reliable vision-based bronchoscopy localization. The introduced synthetic benchmark dataset provides a valuable resource for standardized evaluation on bronchoscopy navigation.}

\keywords{Bronchoscopy, Endoscopic localization, Vision Foundation Models, Computer vision for surgical navigation}



\maketitle

\section{Introduction}\label{intro}
Bronchoscopy is a common routine procedure for facilitating the diagnosis and treatment of various pulmonary conditions. It enables direct visualization of the interior of the lungs and air passages through the insertion of a thin, flexible endoscope equipped with a light and a camera. Robotic-assisted bronchoscopy systems, such as the MONARCH™ Platform, provide physicians with improved maneuverability and stability, allowing access to smaller and more peripheral lung lesions previously difficult to reach. These systems use advanced imaging and navigation technologies, which facilitate precise navigation to target lesions, improve diagnostic performance, and allow simultaneous staging and treatment within a single procedure~\cite{bronchoscopy}.

Despite advances in robotics, accurately localizing the bronchoscope tip within the complex bronchial tree remains challenging, particularly when reaching peripheral lung lesions. Maintaining alignment between the live bronchoscopic view and the pre-operative CT is critical for accurate targeting. Electromagnetic (EM) tracking provides real-time navigation by estimating the bronchoscope pose relative to the CT coordinate frame after initial patient registration.
However, respiratory motion, patient repositioning, and tissue deformation cause the live view to drift from the CT-aligned EM pose, a phenomenon known as CT-to-body divergence~\cite{pritchett2020virtual}.
Since the CT volume is static, this divergence reflects accumulated error in the EM-estimated pose rather than anatomical change. Vision-based methods correct this drift by aligning the live endoscopic scene with CT-derived anatomy, offering appearance-based refinements that maintain consistent CT registration and improve localization accuracy.
%Vision-based methods compensate for this drift by analyzing the live endoscopic scene and aligning it with CT-derived anatomy, providing appearance-based corrections to the EM pose.
%Compared with EM tracking alone, the combined EM and vision approach maintains consistent registration with CT and improves localization accuracy throughout the procedure. 
Approaches include CNN-based pose regression~\cite{offsetnet,airwaynet,zhao2019generative,shen2019context}, generative domain adaptation~\cite{banach2021visually,zhao2019generative}, depth-based estimation~\cite{tian2024dd,shen2019context,ozyoruk2020endoslam,visentini2017deep}, and feature-based correspondence tracking~\cite{teufel2024oneslam,deng2023feature,rodriguez2022tracking,elvira2024cudasift}.
All aim to integrate visual information with CT anatomy for accurate, real-time bronchoscopy navigation.
However, models trained on synthetic or phantom data often fail to generalize to real intra-operative images due to the domain gap between simulated and real scenes, while patient-specific anatomical variability further challenges robustness and leads to localization errors. In addition, there is currently no publicly available bronchoscopy navigation dataset that provides paired CT scans, endoscopic images, depth maps, and ground-truth camera poses, making standardized evaluation and fair comparison across methods difficult. These limitations motivate the development of a more generalizable and anatomically informed framework, together with a comprehensive benchmark, to establish a solid foundation for reliable and reproducible bronchoscopy navigation.

We propose \ours a novel vision-based pose optimization pipeline that establishes a new paradigm for bronchoscopy localization.
The framework unifies domain-invariant visual representation learning, iterative pose optimization, and differentiable rendering-based refinement to achieve robust frame-wise 2D–3D registration between intra-operative endoscopic views and pre-operative CT anatomy.
% Trained entirely on synthetic data, it effectively bridges the domain gap and enables accurate, generalizable localization without any case-specific adaptation.
%We present a vision-based pose optimization framework that performs frame-wise 2D–3D registration by aligning each intra-operative endoscopic scene with its anatomically corresponding location in the pre-operative CT. %Unlike prior work that rely on explicit domain adaptation~\cite{zhao2019generative,banach2021visually}, which often requires case-specific tuning and fails to generalize to unseen distributions, our method employs a fine-tuned representation encoder that is both modality- and domain-invariant, enabling seamless integration of real endoscopic RGB images and synthetic rendered depth maps for iterative pose refinement.
% Unlike prior methods that rely on explicit domain adaptation~\cite{zhao2019generative,banach2021visually}, which often require case-specific tuning and fail to generalize to unseen data, our approach takes a different direction.
We employ a fine-tuned representation encoder that learns modality- and domain-invariant features, allowing real endoscopic RGB images and synthetic rendered depth maps to be seamlessly integrated for cross-modal alignment.
An iterative pose optimization network then estimates the bronchoscope’s 6 DOF pose through successive updates, achieving precise frame-wise 2D–3D registration with the CT anatomy. %Trained exclusively on synthetic data, the model exhibits strong generalization to real patient cases without any additional domain adaptation. %Additionally, a differentiable rendering-based~\cite{ding2022carts,ding2023rethinking} feedback loop iteratively refines pose predictions, using each optimized pose to reinitialize the model for subsequent updates, leading to progressively improved 2D–3D registration accuracy, overcoming key limitations in existing vision-based approaches.
We further introduce a differentiable rendering-based~\cite{ding2022carts,ding2023rethinking} feedback loop that links visual appearance with geometric consistency.
%At each iteration, the optimized pose is used to render a depth view of the CT mesh, which is compared to the inferred depth of the real bronchoscopic image to assess alignment quality.
%The resulting feedback refines the pose prediction and reinitializes the model for the next update, forming a closed-loop optimization that progressively improves 2D–3D registration accuracy.
%This iterative refinement enables the model to correct residual misalignments and overcome limitations of one-shot vision-based approaches.
%Furthermore, 

To enable reproducibility, we construct the first public synthetic benchmark dataset for bronchoscopy navigation. Given the privacy constraints and lack of real patient data with paired CT and ground-truth camera poses, this benchmark offers a controlled, reproducible, and scalable platform for standardized evaluation and comparison.
Quantitative experiments on the proposed benchmark validate the effectiveness of our framework, achieving an average relative pose error of $2.65$ mm in translation, $0.19$ rad in rotation, and a $96\%$ success rate, demonstrating superior accuracy and robustness. Qualitative experiments on real patient cases further confirm strong generalization and reliable performance across real clinical scenarios.
% The major contributions of this work are as follows:
% \begin{itemize}
%     \item A novel vision-based pose optimization framework that unifies domain-invariant representation learning, iterative pose refinement, and differentiable rendering into a closed-loop formulation for accurate frame-wise 2D–3D registration in bronchoscopy navigation.
%     \item The first synthetic benchmark dataset for bronchoscopy navigation, constructed with paired CT meshes, rendered images, depth maps, and ground-truth camera poses, providing a reproducible and standardized platform for quantitative evaluation.
%     \item Comprehensive experimental validation on both synthetic and real patient data, demonstrating superior accuracy, robustness, and generalization compared with existing vision-based localization approaches.
% \end{itemize}

\section{Methods}\label{method}
\subsection{Overview}
\begin{figure}[!t]
\centering
\includegraphics[width=\textwidth]{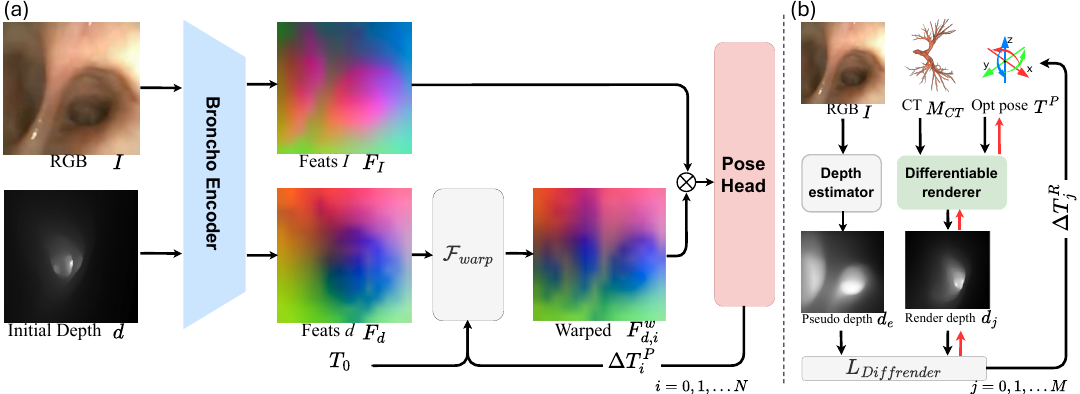}
\caption{The proposed vision-based pose optimization pipeline. (a) The network performs frame-wise 2D–3D registration between a real bronchoscopic RGB image $I$ and a rendered CT-derived depth map $d$.
Given the initial camera pose $T_{0}$, $F_d$ is warped to $F_I$ space, and both features are fused then processed by a pose head that predicts pose increments $\Delta T^{P}_{i}$, progressively aligning the two views. (b) The optimized pose $T^{P} = T_{0} \prod_{i=1}^{N} \Delta T^{P}_{i}$ is used to render a depth map $d_{0}$ from the CT mesh, while a depth estimator infers a pseudo depth map $d_e$ from the live scene $I$.
The rendering loss $L_{Diffrender}(d_e, d_j)$ enforces depth consistency between the two, refining pose estimates $T^R =T^{P} \prod_{j=1}^{M} \Delta T^{R}_{j}$ in a closed-loop manner for improved registration accuracy.
The red arrow denotes the backpropagation path of the differentiable rendering loop.}
\label{fig:overview}
\end{figure}

\ours consists of a representation encoder, a pose optimization network (Fig.~\ref{fig:overview} (a)), and a differentiable rendering-based pose refiner (Fig.~\ref{fig:overview}(b)). 
Our fine-tuned DINO~\cite{simeoni2025dinov3} Broncho Encoder (Sec.~\ref{encoder}) acts as a domain- and modality-invariant feature extractor.
Given a paired RGB image $I \in \mathbb{R}^{H\times W \times 3}$ and rendered depth map $d \in \mathbb{R}^{H\times W \times 1}$, it produces feature embeddings $F_I, F_d \in \mathbb{R}^{\frac{H}{s}\times \frac{W}{s}\times D}$ that are used to iteratively update the camera pose $T \in \SE(3)$ through feature alignment in an end-to-end manner. The differentiable rendering refiner (Sec.~\ref{diffrender}) further refines the estimated pose by enforcing view consistency via image similarity.
The overall system operates in a nested control loop: the inner loop runs the pose optimization network for several iterations to update the camera pose, followed by the differentiable rendering refiner for additional optimization steps.
The refined pose is then used to reinitialize the pose optimization network in the outer loop, forming a closed feedback cycle that progressively improves registration accuracy and ensures stable convergence.

%We present a vision-based intra-operative bronchoscope pose optimization pipeline, as illustrated in Fig.~\ref{fig:overview}. 
% The pipeline (Fig.~\ref{fig:overview}) is built upon a fine-tuned DINO~\cite{simeoni2025dinov3} encoder (Sec.~\ref{encoder}), which serves as a domain- and modality-invariant feature extractor bridging real endoscopic RGB images and synthetic rendered depth maps. Using this encoder, we design a relative pose estimation framework (Sec.~\ref{pose}) that takes a RGB image $I \in \mathbb{R}^{H\times W \times 3}$ and rendered depth map $d \in \mathbb{R}^{H\times W \times 1}$ pair as input and iteratively updates the camera pose $T \in \SE(3)$ in an end-to-end manner to align their feature embeddings $F \in \mathbb{R}^{\frac{H}{s}\times \frac{W}{s} \times D}$. The estimated pose is further refined by a differentiable rendering module (Sec.~\ref{diffrender}), which enforces view consistency through image similarity. The refined pose and corresponding rendered depth are then fed back to reinitialize the pose model, forming a closed-loop optimization process that progressively improves registration accuracy.

\subsection{Modality- and Domain-invariant Encoder Fine-tuning}\label{encoder}
We fine-tune a DINO~\cite{simeoni2025dinov3} Vision Transformer (ViT-S/16) as the Broncho Encoder to bridge the domain and modality gap between synthetic rendered depth maps and real bronchoscopic RGB images.
The encoder is structured as a Siamese network with shared weights, where each branch processes randomly sampled variants of the same anatomical view—such as rendered depth, style-transferred RGB, or their augmented forms.
An InfoNCE loss~\cite{oord2018representation} is applied between the two feature embeddings to enforce modality and domain invariance, pulling together representations from the same view while pushing apart those from different views.

Synthetic training data are generated from pre-operative CT airway models with known camera poses, while realistic bronchoscopic images are produced via CycleGAN~\cite{zhu2020unpairedimagetoimagetranslationusing} translation trained on real patient videos.
To improve robustness, we apply strong photometric and geometric augmentations, including random brightness, blur, elastic deformation, and coarse dropout, simulating variations in illumination, occlusion, and motion.
Through this process, the encoder learns a shared, geometry-aware representation space that aligns features across modalities, providing a robust foundation for the subsequent pose optimization framework (Sec.~\ref{pose}) to perform cross-modal 2D–3D registration without explicit domain adaptation.

\subsection{Pose Optimization Framework}\label{pose}
Building upon the Broncho Encoder, we design a vision-based pose optimization framework (Fig.~\ref{fig:overview} (a)) that performs frame-wise 2D–3D registration between intra-operative endoscopic RGB images and pre-operative CT-derived renderings.
Given an input pair consisting of a real bronchoscopic image $I$ and a rendered depth map $d$, both are passed through the Broncho Encoder to extract aligned feature embeddings denoted as $F_I$ and $F_d$. Using an initial relative camera pose $T_{init}$, the depth-map feature $F_d$ is spatially aligned to the RGB feature space via differentiable warping:
\begin{align}
    p_{I} \sim \pi\!\left(T_{\text{init}} \, \pi^{-1}(p_d, d(p_d)) \right), \quad  F_d^{w}(p_I) = F_d(p_d)
\end{align}
where $\pi(\cdot)$ and $\pi^{-1}(\cdot)$ denote the camera projection and back-projection using depth $d$ and $p_I\, ,p_d$ denotes pixel coordinates in image and depth map respectively.

The aligned features $F_I$ and $F_d^{w}$ are then fused using a Cross-Patch Fuser, a stack of LoFTR-like~\cite{sun2021loftr} cross-attention blocks that jointly attend to local correspondences and global context.
The fused tokens form compact pose descriptors, which are processed by two
additional self-attention layers followed by a lightweight linear layer to predict the relative transformation $\Delta \boldsymbol{\xi} \in \mathfrak{se}(3)$. The camera pose is iteratively updated in the Lie algebra formulation $\hat{\mathbf{T}}_{i+1} = \hat{\mathbf{T}}_i\exp(\Delta \boldsymbol{\xi}_i)$, 
% \begin{align}
%     \hat{\mathbf{T}}_{k+1} = \hat{\mathbf{T}}_k\exp(\Delta \boldsymbol{\xi}_k)
% \end{align}
where translation and rotation updates are dynamically scaled by learnable confidence factors.
The optimization is supervised using known camera poses $\mathbf{T} = [\mathbf{R} | \mathbf{t}]$ from synthetic data, with losses defined on both rotation and translation components given predicted pose $\hat{\mathbf{T}} = [\hat{\mathbf{R}} | \hat{\mathbf{t}}]$. The overall pose loss is expressed as:
\begin{align}\label{poseloss}
    \mathcal{L}_{\text{pose}}
= \lambda_{t}\underbrace{\arccos\!\left(
\frac{\hat{\mathbf{t}}^\top \mathbf{t}}
{\|\hat{\mathbf{t}}\|\|\mathbf{t}\|}
\right)}_{\text{translation direction}}
	+	\lambda_{rot}\underbrace{\mathcal{L}_{\text{geo}}(\hat{\mathbf{R}}, \mathbf{R})}_{\text{rotation geodesic}}
	+	\lambda_{tm}\underbrace{\|\|\hat{\mathbf{t}}\| - \|\mathbf{t}\|\|_1}_{\text{translation magnitude}}
\end{align}
where $\mathcal{L}_{\text{geo}}(\hat{\mathbf{R}}, \mathbf{R}) = \|\log(\hat{\mathbf{R}}^\top \mathbf{R})\|_2$ measures the geodesic distance on $\mathbb{SO}(3)$, and $\lambda_t \, \lambda_{rot} \,\text{and} \, \lambda_{tm}$ balance the magnitude for each term.

In addition to the pose supervision, we apply a rendering consistency loss to further constrain geometric alignment by encouraging fine-grained depth alignment during optimization.
Given the rendered depth map $\hat{d}$ obtained from the predicted pose and the ground-truth depth $d$ rendered from the same viewpoint, we define the rendering loss $\mathcal{L}_{render}$ using the multi-scale structural similarity (MS-SSIM~\cite{msssim}) metric.
% \begin{align}\label{renderloss}
%     \mathcal{L}_{\text{render}}
% = 1 - \text{MS-SSIM}(normalize(\hat{d}), normalize(d))
% \end{align}
The total training objective combines the pose and rendering losses within a sequence loss formulation that supervises all iterative pose updates.
Specifically, the loss accumulated over N refinement steps is defined as: $\mathcal{L}_{\text{total}}
= \sum_{i=1}^{N} 0.6^{N-i} \big( \mathcal{L}_{\text{pose}}^{(i)} + \lambda_r \mathcal{L}_{\text{render}}^{(i)} \big)$,
% \begin{align}\label{totalloss}
%     \mathcal{L}_{\text{total}}
% = \sum_{i=1}^{N} 0.6^{N-i} \big( \mathcal{L}_{\text{pose}}^{(i)} + \lambda_r \mathcal{L}_{\text{render}}^{(i)} \big)
% \end{align}
where $\lambda_{r}$ balances the relative contribution of the rendering loss.

% At inference, the framework generalizes to unseen synthetic data and real patient data without requiring retraining or explicit domain adaptation, enabling direct 2D–3D registration between real bronchoscopic frames and pre-operative CT anatomy.

\subsection{Differentiable Rendering-based Pose Refinement}\label{diffrender}
To further enhance localization accuracy and enforce geometric consistency, we introduce a differentiable rendering-based refinement module (Fig.~\ref{fig:overview}(b)) following the pose optimization stage.
Given a predicted camera pose $T^P$ and the corresponding pre-operative CT-derived airway mesh $ M_{CT}$, a differentiable renderer~\cite{Laine2020diffrast} synthesizes a depth image $d_0 = \mathcal{R}(T^P, M_{CT})$. The rendered depth map $d_0$ is then compared against the pseudo depth map $d_e$ inferred from the intra-operative endoscope image using an off-the-shelf depth estimator~\cite{paruchuri2024leveragingnearfieldlightingmonocular}. 
The rendering consistency is enforced by $\mathcal{L}_{\text{render}}$.
To prevent pose updates that move the virtual camera outside the airway lumen, we additionally impose a Signed Distance Field (SDF) constraint.
Let $s$ denote the signed distance field (SDF) value at the camera center in 3D space, where $s < 0 $ indicates that the camera lies inside the airway lumen and $s > 0$ indicates that it is outside the airway surface.
The SDF loss is defined as:
\begin{align}\label{sdfloss}
    \mathcal{L}_{\text{sdf}}
= w_\text{in} \, \text{softplus}\!\big(\gamma(s + \tau)\big)^2
	+	w_\text{near}\, \big[-\log(s_\text{neg}/\tau)\big]_+
	+	w_\text{out} \, \text{ReLU}(s)^2
\end{align}
where the first term penalizes points approaching the airway wall,
the second applies a logarithmic barrier within a narrow boundary layer near the surface,
and the third imposes a quadratic penalty for points outside the airway.
Here, $\tau$ defines the safety margin near the airway wall, $w$ controls the relative penalty strengths, and $\gamma$ adjusts the sharpness of the soft boundary. The final objective combines all components as $\mathcal{L}_{\text{diffrender}} = \lambda_r\mathcal{L}_{\text{render}} + \lambda_s\mathcal{L}_{\text{sdf}}$, 
% \begin{align}\label{diffrenderloss}
%     \mathcal{L}_{\text{diffrender}} = \lambda_r\mathcal{L}_{\text{render}} + \lambda_s\mathcal{L}_{\text{sdf}}
% \end{align}
where $\lambda_r$ and $\lambda_s$ weight the rendering and SDF terms.

\section{Experiments}\label{exp}
\subsection{Datasets}
% \subsubsection{Synthetic Dataset}\label{syndata}
\noindent\textbf{Synthetic Dataset}
%\textbf{Training Set} 
We generate a synthetic dataset for Broncho Encoder fine-tuning and pose optimization network training using pre-operative CT scans from our collected patient dataset.
The dataset is designed to simulate bronchoscope motion with geometric and anatomical consistency.
Camera poses are uniformly sampled along airway centerlines, with the optical axis aligned to the local tangent direction.
To introduce variability, random perturbations are applied within a $\pm 10$ mm translation range and $\pm 0.44$ rad rotation range, reflecting the typical magnitude of bronchoscope motion and pose drift observed during clinical navigation~
\cite{borrego2023bronchopose}, while maintaining sufficient co-visible regions between paired views for stable geometric correspondence.
Depth and RGB images are rendered from the pre-operative CT mesh, and a CycleGAN~\cite{zhu2017unpaired} model is trained to synthesize realistic bronchoscopic textures for domain adaptation.
In total, the dataset contains $481,650$ rendered images for Broncho Encoder fine-tuning and $60,000$ paired samples for pose model training.

%\textbf{Evaluation Set} 
To establish a benchmark dataset for evaluation, we adopt a generation strategy consistent with that used for the training set.
Instead of our collected patient CTs—which remain confidential due to data privacy restrictions—we utilize $15$ publicly available CT cases from the AeroPath dataset~\cite{stoverud2023aeropath} to generate unseen data pairs.
The benchmark pairs are further categorized into three difficulty levels based on the pose loss in Eq.~\ref{poseloss}:
easy ($< 0.4$), medium ($0.4\text{–}0.8$), and hard ($0.8\text{–}1.6$).
This stratification enables fine-grained evaluation of model performance under varying pose disparities and geometric complexities.
In total, the benchmark dataset contains $10,715$ image pairs, providing a standardized and publicly reproducible platform for quantitative evaluation of bronchoscopy pose estimation methods. Detailed dataset statistics, structure, and representative examples are provided in the Appendix A.

% \subsubsection{Real Patient Dataset}
\noindent\textbf{Real Patient Datset}
We collected a video dataset using the MONARCH™ Platform, a robotic-assisted bronchoscopy system indicated for diagnostic and therapeutic lung procedures. The dataset includes video of endoscopic insertion and navigation from the trachea to the peripheral airways. Frames containing surgical instruments were removed to
ensure a clear view of the airways for the development of vision-based localization pipelines. The dataset includes $8$ CTs and $8$ human bronchoscopy videos, with $20,555$ frames in total. Videos were collected at $30$ fps with a resolution of $200\ px\times 200\ px$. Each procedure has an accompanying pre-operative CT scan of the airway anatomy, along with the segmented airway tree. Airway centerlines were generated from the airway tree segmentation model. 6-DoF bronchoscope poses are recorded using an EM sensor and registered to pre-operative CT coordinates. These poses are synchronized with image frames via timestamps. %In real patient cases, intra-operative deformation causes misalignment between the poses and pre-operative CT.

\subsection{Implementation Details}
The framework is implemented in PyTorch and trained on an NVIDIA H100 GPU, while experiments are performed on a TITAN GPU. During training, all input depth and disparity maps are normalized and resized to $224\times224$ pixels. The comprehensive hyper-parameters are provided in the Appendix C. %The BronchoEncoder is trained using the AdamW optimizer with a learning rate of $1\times10^{-4}$ and a OneCycleLR scheduler for $300$ epochs. The pose optimization stage adopts the same optimizer and learning rate schedule for $120$ epochs. In the pose loss (Eq.\ref{poseloss}), the weights are set to $\lambda_{r}=0.5$, $\lambda_{t}=1$, $\lambda_{rot}=1$, and $\lambda_{tm}=100$. For the SDF loss (Eq.\ref{sdfloss}), we set $w_{\text{in}}=w_{\text{out}}=w_{\text{near}}=1$. The pose optimizer and differentiable rendering refiner each run for $i=3$ inner iterations, and the overall optimization pipeline is repeated for $k=3$ outer iterations. In the differentiable rendering refinement (Eq.~\ref{diffrenderloss}), we use $\lambda_{r}=1$ and $\lambda_{s}=0.1$.

\subsection{Evaluation Metrics}
\noindent\textbf{Similarity Metrics:}
Since real patient cases lack ground-truth camera poses, it is not feasible to directly quantify pose accuracy for these data.
Therefore, in addition to standard pose metrics evaluated on the synthetic dataset, we rely on image similarity-based metrics as proxies for alignment quality between optimized rendered and observed views.
These similarity measures provide indirect yet reliable indicators of pose accuracy by evaluating geometric and structural consistency between views.
Normal Correlation (NC): quantifies local surface orientation consistency by measuring cosine similarity between surface normals.
Depth Similarity (DS): both predicted and rendered depth maps are mean–absolute–deviation normalized (MADNorm), after which cosine similarity is computed between them to assess overall depth agreement.
Scale-Invariant Structural Alignment (SI)~\cite{eigen2014depth}: quantifies global geometric consistency independent of absolute scale. 

\noindent\textbf{Pose Metrics:}
Pose accuracy is assessed using the translational $L_2$ distance and the rotational geodesic distance between the predicted pose and the ground truth.

\begin{table*}[!t]
\caption{Quantitative comparison across difficulty levels between OffsetNet~\cite{offsetnet} and the proposed BronchOpt. The initial camera poses (Init) from the robotic system show increasing difficulty across levels, with mean translation/rotation distances of
$1.94 \pm 2.03$ mm / $0.14 \pm 0.11$ rad for Easy,
$4.29 \pm 2.68$ mm / $0.34 \pm 0.15$ rad for Medium, and
$6.96 \pm 3.09$ mm / $0.49 \pm 0.25$ rad for Hard cases. The reported metrics are computed only over successful cases.}
\centering
\label{tab:main}
\scalebox{0.68}{
\begin{tabular}{c|c|c|c|c|c|c|c}
\toprule
\multirow{2}{*}{\textbf{Levels}} & \multirow{2}{*}{\textbf{Models}} & \multirow{2}{*}{\textbf{DS}$\uparrow$} & \multirow{2}{*}{\textbf{NC}$\uparrow$} & \multirow{2}{*}{\textbf{SI}$\downarrow$} & \textbf{Trans. Err}$\downarrow$ & \textbf{Rot. Err}$\downarrow$ \multirow{2}{*} &{\textbf{Success Rate}}\\
&  &  &  &  & (mm) & (rad) & $\%$ \\
\midrule
\multirow{2}{*}{\textbf{Easy}}   & OffsetNet~\cite{offsetnet} & 0.76 $\pm$ 0.16 &0.54 $\pm$ 0.20  & 0.58 $\pm$ 1.12 & 4.60 $\pm$ 2.52& 0.15 $\pm$ 0.10 & 37.1 \\
& \textbf{\ours} & \textbf{0.94 $\pm$ 0.07} & \textbf{0.74 $\pm$ 0.11} & \textbf{0.08 $\pm$ 0.28} & \textbf{1.81 $\pm$ 2.09} & \textbf{0.13 $\pm$ 0.09} & \textbf{94.7} \\
\multirow{2}{*}{\textbf{Medium}}   & OffsetNet~\cite{offsetnet} & 0.68 $\pm$ 0.16 & 0.44 $\pm$ 0.19 & 0.75 $\pm$ 1.01 & 6.17 $\pm$ 3.13 & 0.34 $\pm$ 0.15 & 45.5 \\
& \textbf{\ours} & \textbf{0.92 $\pm$ 0.09} & \textbf{0.69 $\pm$ 0.13} & \textbf{0.11 $\pm$ 0.39} & \textbf{2.98 $\pm$ 3.10} & \textbf{0.21 $\pm$ 0.14} & \textbf{96.8} \\
\multirow{2}{*}{\textbf{Hard}}   & OffsetNet~\cite{offsetnet} & 0.63 $\pm$ 0.16 & 0.37 $\pm$ 0.19 & 0.99 $\pm$ 1.15 & 8.09 $\pm$ 3.42 & 0.47 $\pm$ 0.24 & 56.5  \\
& \textbf{\ours} & \textbf{0.90 $\pm$ 0.11} & \textbf{0.65 $\pm$ 0.16} & \textbf{0.15 $\pm$ 0.32} & \textbf{4.53 $\pm$ 3.93} & \textbf{0.31 $\pm$ 0.24} & \textbf{98.3} \\
\multirow{2}{*}{\textbf{Average}}   & OffsetNet~\cite{offsetnet} & 0.70 $\pm$ 0.17 & 0.447 $\pm$ 0.20 & 0.72 $\pm$ 1.09 & 5.87 $\pm$ 3.22 & 0.28 $\pm$  0.20 & 43.0 \\
& \textbf{\ours} & \textbf{0.93 $\pm$ 0.09} & \textbf{0.71 $\pm$ 0.14} & \textbf{0.10 $\pm$ 0.33} & \textbf{2.65 $\pm$  2.96} & \textbf{0.19 $\pm$ 0.15} & \textbf{96.0}\\
\bottomrule
\end{tabular}
}
\footnotetext[1]{0}
\footnotetext[2]{0}
\end{table*}

\subsection{Quantitative Evaluation on Synthetic Dataset}
We conduct quantitative experiments on the synthetic benchmark dataset to evaluate both pose accuracy and geometric alignment.
Our method is compared against OffsetNet~\cite{offsetnet}. For a fair comparison, both methods are trained on the same synthetic data and tested on the benchmark dataset containing unseen CT cases.
In addition to quantitative errors, we report the success rate, defined as the proportion of cases where the estimated camera center remains within the airway lumen.
% The initial camera poses (Init) from the robotic system show increasing difficulty across levels, with mean translation/rotation distances of
% $1.94 \pm 2.03$ mm / $0.14 \pm 0.11$ rad for Easy,
% $4.29 \pm 2.68$ mm / $0.34 \pm 0.15$ rad for Medium, and
% $6.96 \pm 3.09$ mm / $0.49 \pm 0.25$ rad for Hard cases.
Overall, our method achieves consistently lower translation and rotation errors and higher similarity scores than OffsetNet~\cite{offsetnet} across all difficulty levels (Tab.~\ref{tab:main}).
The improvement in Normal Correlation and Depth Similarity demonstrates better geometric consistency between rendered and predicted views, while the higher success rate indicates robust, anatomically valid pose estimation with rare drift or collapse outside the airway lumen.

Despite the overall improvement, some alignment failures occur in higher-generation bronchi, where the lumens are narrow and irregular.
These cases likely result from the encoder’s reduced feature sensitivity in regions derived from low-resolution CT meshes, where limited geometric detail produces less informative depth and texture cues. As a result, correspondence between rendered and real views becomes less stable, reducing registration accuracy in these challenging regions.

Separately, while the proposed benchmark provides a valuable and standardized platform for evaluating bronchoscopy localization, its controlled synthetic environment cannot fully reproduce the lighting variations and anatomical deformations present in real clinical cases.
Future extensions aim to better capture the visual and geometric characteristics of real bronchoscopic videos, thereby enhancing the benchmark’s clinical relevance.

\subsection{Qualitative Evaluation on Real Patient Dataset}
\begin{figure}[!t]
\centering
\includegraphics[width=\textwidth]{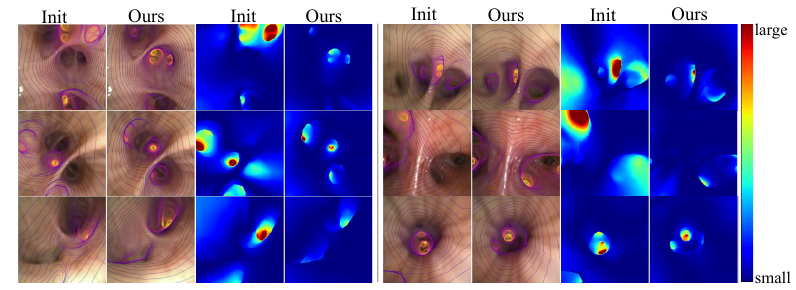}
\caption{Qualitative alignment and depth error visualization on real patient data. First two columns: Contour overlays illustrate the geometric correspondence between rendered CT views and real bronchoscopic images.
Initially, the contours show noticeable misalignment with the real scene, but after applying our optimization, they align closely, indicating improved pose estimation accuracy.
Last two columns: depth error maps before (Init) and after optimization (Ours).}
\label{fig:errormap}
\end{figure}

We evaluate the proposed method on real bronchoscopic videos to assess its cross-domain generalization capability, where ground-truth camera poses are unavailable.
In these experiments, the initial pose (Init) refers to the camera pose estimated from the robotic system’s onboard sensors, which serves as the starting point for our optimization pipeline. We did not include OffsetNet~\cite{offsetnet} for qualitative comparison because it failed to generalize to real patient data.
Using both the initial and the optimized poses, we render the CT-based depth maps and corresponding contours, which are then overlaid on the real bronchoscopic images to intuitively illustrate geometric alignment.
Additionally, we compute error maps based on the scale-invariant depth difference between the rendered depth and the inferred depth~\cite{paruchuri2024leveragingnearfieldlightingmonocular} from the real RGB image.
Cooler colors represent lower geometric discrepancy, while warmer regions highlight residual misalignment.
Fig.~\ref{fig:errormap} presents representative qualitative results.
Compared with the initial robotic readings, our optimized results show noticeably improved contour alignment and reduced high-error regions, demonstrating that the proposed method effectively refines pose estimates and achieves anatomically consistent 2D–3D registration.

In some real patient cases, alignment errors are still observed.
The differentiable rendering module can struggle when the inferred depth map is ambiguous, particularly under strong illumination changes that cause the depth model~\cite{paruchuri2024leveragingnearfieldlightingmonocular} to misinterpret bright regions as deep lumens.
Additionally, pronounced anatomical deformation and CT-to-body divergence may introduce discrepancies between the live bronchoscopic view and the pre-operative CT geometry, leading to localized misalignment.
To address these issues, we plan to train a dedicated bronchoscopic depth estimation model using domain-adapted data and incorporate deformation-aware alignment strategies to improve robustness under challenging lighting and anatomical conditions. Additional qualitative results and visual examples are provided in the Appendix B.
% Despite the overall improvement, failures can occur in certain real cases.
% The differentiable rendering module may fail when the inferred depth is ambiguous, often due to strong illumination variations that cause the depth model~\cite{paruchuri2024leveragingnearfieldlightingmonocular} to misinterpret bright regions as deep lumens.
% Severe anatomical deformation and CT-to-body divergence can also lead to mismatches between intra-operative views and pre-operative CT geometry, resulting in local misalignment.

\begin{table*}[!t]
\caption{Ablation study showing the effect of each module—encoder (E), pose optimizer (P), and differentiable rendering refiner (D)—on geometric similarity, pose accuracy and sucess rate.}
\centering
\label{tab:ablate}
\scalebox{0.58}{
\begin{tabular}{l|ccc|cccccc}
\toprule
\multirow{2}{*}{\textbf{Methods}} & \multicolumn{3}{|c|}{\textbf{Modules}} & \multirow{2}{*}{\textbf{DS}$\uparrow$} & \multirow{2}{*}{\textbf{NC}$\uparrow$} & \multirow{2}{*}{\textbf{SI}$\downarrow$} & \textbf{Trans. Err}$\downarrow$ & \textbf{Rot. Err}$\downarrow$ &
\textbf{Success Rate} \\
\cmidrule{2-4}
 & \textbf{E} & \textbf{P} & \textbf{D} &  &  &  & (mm) & (rad) & \% \\
\midrule
DINO + MLP & & & & 0.70 $\pm$ 0.15 & 0.46  $\pm$ 0.17 & 0.69  $\pm$ 0.94  & 6.96   $\pm$ 3.47 & 0.31   $\pm$ 0.16 & 45.7 \\
DINO (w/o PoseOpt.) &  &  & \checkmark & 0.74  $\pm$ 0.15 & 0.48  $\pm$ 0.17 & 0.50  $\pm$ 0.78 & 6.32  $\pm$ 3.38 & 0.31  $\pm$ 0.16 & 69.4 \\
DINO (w/o DiffRend.) &  & \checkmark &  & 0.80  $\pm$ 0.13 & 0.55  $\pm$ 0.17  & 0.33  $\pm$ 0.52 & 3.75  $\pm$ 2.79 & 0.27  $\pm$ 0.17 & 90.0 \\
DINO (w PoseOpt. \& DiffRend.) &  & \checkmark & \checkmark & 0.83  $\pm$ 0.12 & 0.58  $\pm$ 0.17 & 0.25  $\pm$ 0.45 & 3.56  $\pm$ 2.70 & 0.27  $\pm$ 0.17 & 93.5 \\
BronchoEncoder + MLP & \checkmark &  &  & 0.85  $\pm$ 0.13 & 0.59  $\pm$ 0.16 & 0.25 $\pm$ 0.61 & 5.50  $\pm$ 4.83 & 0.26  $\pm$ 0.18 & 68.7 \\
\ours (w/o PoseOpt.) & \checkmark &  & \checkmark & 0.86  $\pm$ 0.12 & 0.60  $\pm$ 0.16 & 0.20  $\pm$ 0.66 & 5.03  $\pm$ 4.24 & 0.25  $\pm$ 0.17 & 85.5 \\
\ours (w/o DiffRend.) & \checkmark & \checkmark &  & 0.92  $\pm$ 0.09 & 0.71  $\pm$ 0.13 & 0.11  $\pm$ 0.33 & 2.87  $\pm$ 3.28 & 0.19  $\pm$ 0.16 & 93.0\\
\textbf{\ours (full)} & \checkmark & \checkmark & \checkmark & \textbf{0.93   $\pm$ 0.09} & \textbf{0.71  $\pm$ 0.13} & \textbf{0.10  $\pm$ 0.33} & \textbf{2.65  $\pm$ 2.96} & \textbf{0.19  $\pm$ 0.15} & \textbf{96.0} \\
\bottomrule
\end{tabular}
}
\end{table*}

\subsection{Ablation Study}
To evaluate the contribution of each module in our framework, we conduct a series of ablation experiments summarized in Table \ref{tab:ablate}.
We denote E, P, and D as the encoder, pose optimizer, and differentiable rendering refiner modules, respectively. A checkmark under each column denotes the inclusion of the corresponding component. Specifically, E indicates whether the model uses the generic DINO~\cite{simeoni2025dinov3} encoder or our fine-tuned Broncho Encoder; P denotes whether the pose regression is performed by a simple MLP or by our iterative pose optimizer (PoseOpt.); and D determines whether the differentiable rendering (DiffRend.)–based refinement is applied.

% Replacing the DINO encoder with the proposed Broncho Encoder consistently improves geometric similarity and pose estimation accuracy. When only an MLP head is used, the Broncho Encoder increases DS from $0.70$ to $0.85$ and NC from $0.46$ to $0.59$, highlighting the benefit of domain-specific representation learning.

% Introducing the Pose Optimizer further enhances performance. Compared with BronchOpt (w/o PoseOpt.), the full model reduces translational and rotational errors from $5.03$ mm / $0.25$ rad to $2.65$ mm / $0.19$ rad, while the success rate rises from $85.5 \%$ to $96.0 \%$. These results confirm that iterative optimization effectively refines the predicted camera pose beyond what a single-shot regression can achieve.

% The differentiable rendering module also contributes to better structural alignment by allowing small pose adjustments guided by the depth alignment loss, effectively reducing residual discrepancies that are difficult to correct through feature-based regression alone. When integrated with the pose optimizer, DiffRend. improves NC from $0.55$ to $0.58$ and lowers SI from $0.33$ to $0.25$ on the DINO backbone. 
\noindent\textbf{Broncho Encoder}
Replacing the generic DINO encoder with the proposed Broncho Encoder consistently improves geometric similarity and pose estimation accuracy.
When only an MLP head is used, the Broncho Encoder increases DS from $0.70$ to $0.85$ and NC from $0.46$ to $0.59$, demonstrating the effectiveness of domain-specific representation learning for bronchoscopy.

\noindent\textbf{PoseOpt}
Introducing the PoseOpt. further enhances performance.
Compared with BronchOpt (w/o PoseOpt.), the full model reduces translational and rotational errors from $5.03$ mm / $0.25$ rad to $2.65$ mm / $0.19$ rad, while the success rate increases from $85.5\%$ to $96.0\%$.
These improvements confirm that iterative optimization effectively refines camera poses beyond what single-shot regression can achieve.

\noindent\textbf{DiffRend}
The DiffRend. module further improves structural alignment by enabling fine pose adjustments guided by depth alignment loss, reducing residual discrepancies that feature-based regression alone cannot correct.
When integrated with the PoseOpt., DiffRend. increases NC from $0.55$ to $0.58$ and decreases SI from $0.33$ to $0.25$ on the DINO backbone.
Overall, the full configuration (BronchOpt (full)) achieves the best results across all metrics, demonstrating the complementary benefits of the proposed encoder, optimization, and differentiable rendering components.

\section{Conclusion}\label{conclusion}
% We presented~\ours, a vision-based pose optimization framework for accurate bronchoscope localization via frame-wise 2D–3D registration between intra-operative bronchoscopic images and pre-operative CT anatomy.
\ours integrates a modality- and domain-invariant encoder, an iterative pose optimization network, and a differentiable rendering-based refinement module that jointly ensure robust and geometrically consistent alignment.
Trained entirely on synthetic data, our model demonstrates strong generalization to real patient cases without domain-specific adaptation.
Comprehensive experiments on synthetic benchmarks and real videos confirm its superiority over existing approaches in both pose accuracy and structural alignment. %Although the proposed benchmark provides a valuable and standardized platform for evaluation, it remains challenging for conventional methods while being largely solved by our approach.
%However, its controlled synthetic conditions limit its ability to fully capture the complexity and variability of real clinical scenarios.
Future work will focus on improving robustness under illumination variation, anatomical deformation, and CT-to-body divergence, while extending the benchmark to better capture these real-world factors.
% Overall, this work establishes a scalable and anatomically grounded foundation for reliable, real-time bronchoscopy navigation.
Overall, this work establishes a scalable and anatomically grounded foundation for reliable bronchoscopy navigation, complemented by a comprehensive synthetic benchmark that supports standardized evaluation and future methodological advancement.

\backmatter

% \bmhead{Supplementary information}

% \bmhead{Acknowledgements}
% This work is partially sponsored by Johnson \& Johnson MedTech.
% \section*{Declarations}


\bibliography{sn-bibliography}% common bib file
%%%%%%%%%%%%%%%%%% Appendix
\newpage
\begin{appendices}

\section{Synthetic Benchmark Dataset}\label{secA1}
This section provides additional details of the synthetic benchmark dataset used for evaluation.  
As described in synthetic data generation, the dataset is generated using $15$ publicly available CT cases from the AeroPath dataset~\cite{stoverud2023aeropath} following the same procedure as the training data, but with unseen airway geometries.  
Each case contains rendered RGB–depth pairs with ground-truth camera poses sampled along airway centerlines at 5~mm intervals and further perturbed within $\pm 5$~mm translation and $\pm 25^{\circ}$ rotation ranges.  
All paired samples are organized into three difficulty levels (easy, medium, hard) based on the pose loss range: $<0.4$, $0.4$–$0.8$, and $0.8$–$1.6$.  
The dataset contains a total of 10{,}715 image pairs.

\begin{figure}[htb]
    \centering
    \includegraphics[width=0.5\linewidth]{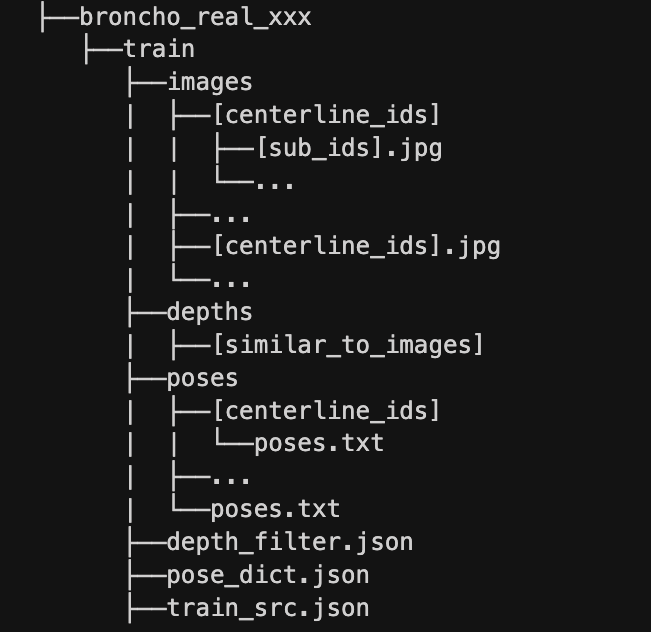}
    \caption{Example data structure for the synthetic benchmark dataset.}
    \label{fig:struct}
\end{figure}

Fig.~\ref{fig:struct} illustrates the folder organization of the dataset, including images, depths, poses, and metadata files used for training and evaluation.  
Each \texttt{poses.txt} file lists the sampled camera transformations corresponding to the rendered images.

\begin{figure}[htb]
    \centering
    \includegraphics[width=\linewidth]{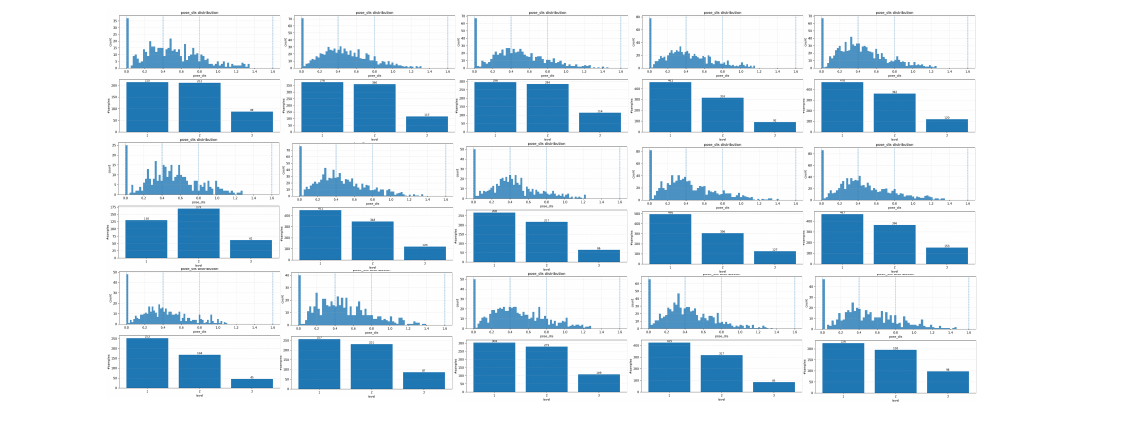}
    \caption{Statistics for each cases.}
    \label{fig:stat}
\end{figure}

Fig.~\ref{fig:stat} summarizes the data statistics for each CT case.  
The histograms show the distribution of pose losses and the number of valid image pairs in each difficulty level.  
This stratified split enables balanced evaluation across a wide range of pose disparities.

\begin{figure}[htb]
\centering
\includegraphics[width=\textwidth]{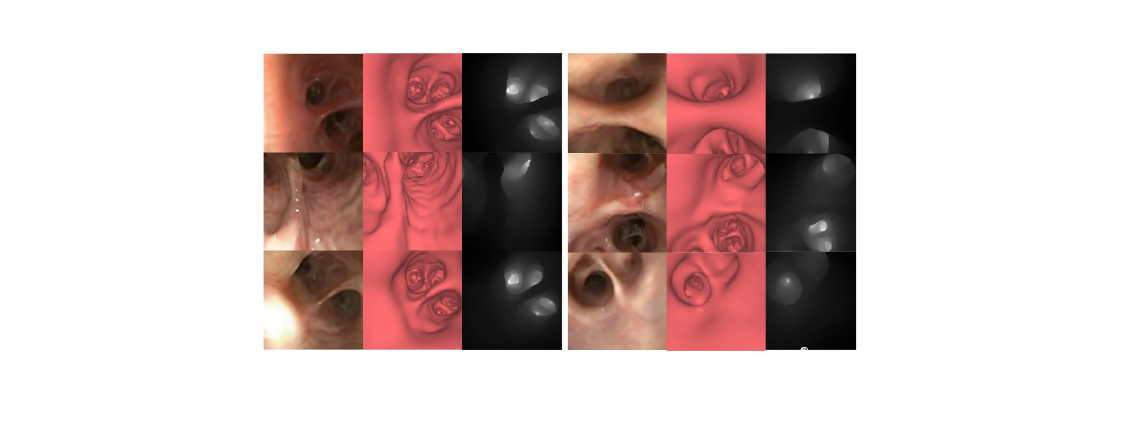}
\caption{ and example image pairs for the synthetic benchmark dataset.}
\label{fig:example}
\end{figure}

Representative examples of RGB–depth pairs from different difficulty levels are shown in Fig.~\ref{fig:example}.


\section{Additional Qualitative Results}
The detailed results of three similarity metrics on all 8 real patient cases are shown in Tab.~\ref{tab:add}.
\begin{table*}[!thb]
\caption{Qualitative results on real patient cases}
\centering
\label{tab:add}
\scalebox{0.8}{
\begin{tabular}{cccc}
\toprule
\textbf{Case ID} & \textbf{DS}$\uparrow$ & \textbf{NC}$\uparrow$ & \textbf{SI}$\downarrow$\\
\midrule
1 &  0.90 $\pm$ 0.05 & 0.75 $\pm$ 0.21 & 0.09 $\pm$ 0.03\\
2 &  0.84 $\pm$ 0.08 &	0.72 $\pm$ 0.42 &	0.14 $\pm$ 0.08 \\
3 &  0.90 $\pm$ 0.05 &	0.75 $\pm$ 0.23 &	 0.09 $\pm$ 0.07 \\
4 &  0.88 $\pm$ 0.07 &	0.74 $\pm$ 0.32 &	0.12 $\pm$ 0.16 \\
5 &  0.83 $\pm$ 0.09 &	0.65 $\pm$ 0.41 &	0.24 $\pm$ 0.44 \\
6 &  0.91 $\pm$ 0.04 &	0.76 $\pm$ 0.24 &	0.12 $\pm$ 0.09  \\
7 &  0.89 $\pm$ 0.09 &	0.66 $\pm$ 0.46 &	0.24 $\pm$ 0.60 \\
8 &  0.81 $\pm$ 0.13 &	0.62 $\pm$ 0.54 &	0.45 $\pm$ 1.48 \\
\bottomrule
\end{tabular}
}
\end{table*}
\section{Hyper-parameters}
The BronchoEncoder is trained using the AdamW optimizer with a learning rate of $1\times10^{-4}$ and a OneCycleLR scheduler for $300$ epochs. The pose optimization stage adopts the same optimizer and learning rate schedule for $120$ epochs. In the pose loss (Eq.\ref{poseloss}), the weights are set to $\lambda_{r}=0.5$, $\lambda_{t}=1$, $\lambda_{rot}=1$, and $\lambda_{tm}=100$. For the SDF loss (Eq.\ref{sdfloss}), we set $w_{\text{in}}=w_{\text{out}}=w_{\text{near}}=1$. The pose optimizer and differentiable rendering refiner each run for $i=3$ inner iterations, and the overall optimization pipeline is repeated for $k=3$ outer iterations. In the differentiable rendering refinement, we use $\lambda_{r}=1$ and $\lambda_{s}=0.1$.

\end{appendices}

\end{document}